\documentclass[a4paper,11pt,twocolumn,twoside]{article}
\usepackage{graphicx}
\usepackage{wallpaper}
\ThisULCornerWallPaper{1}{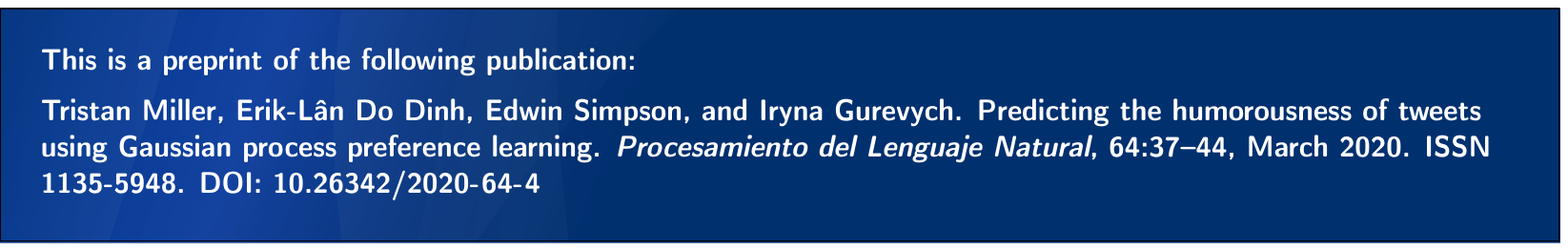}
\usepackage{sepln_en}
\usepackage{fullname}
\usepackage[spanish,es-nosectiondot, es-tabla, es-noindentfirst,canadian]{babel}
\usepackage[utf8]{inputenc}

\usepackage{amssymb} 
\usepackage{textcomp} 
\usepackage{microtype} 
\usepackage{xspace} 
\usepackage{url} 
\usepackage{booktabs} 
\usepackage{xcolor}

\newcommand{\HAHA}[1][2019]{HAHA@IberLEF#1\xspace}
\title{Predicting the Humorousness of Tweets Using Gaussian Process Preference Learning}

\author {\textbf{Tristan Miller$^1$}, \textbf{Erik-Lân Do Dinh$^2$}, \textbf{Edwin Simpson$^2$}, \textbf{Iryna Gurevych$^2$}\\
$^1$Austrian Research Institute for Artificial Intelligence (OFAI)\\
Freyung~6, 1010~Vienna, Austria\\
tristan.miller@ofai.at\\
$^2$Ubiquitous Knowledge Processing Lab (UKP-TUDA)\\
Department of Computer Science, Technische Universität Darmstadt\\
Hochschulstraße~10, 64289~Darmstadt, Germany\\
\url{https://www.ukp.tu-darmstadt.de/}\\
}

\seplntranstitle{Identificando el humor de tuits utilizando el aprendizaje de preferencias basado en procesos gaussianos}

\seplnclave{Humor computacional, humor, preferencias de aprendizaje, proceso gaussiano, GPPL, mejor-peor escala}

\seplnresumen{\foreignlanguage{spanish}{Actualmente la mayoría de los sistemas de procesamiento de humor hacen, en el mejor de los casos, distinciones discretas y granualeres entre lo cómico y lo convencional. Sin embargo, dichos conceptos se conciben mejor en un espectro más amplio. Este artículo presenta un método probabilístico, un modo de preferencias de aprendizaje basadas en un proceso gaussiano~(GPPL), que aprende a clasificar y calificar el humor de textos cortos explotando juicios de preferencia humana y anotaciones lingüísticas generadas en forma automática. Nuestro sistema es similar a uno que previamente había demostrado un buen desempeño en frases en inglés anotadas con anotaciones  humorísticas por pares y lo aplicamos a  la colección de datos en español de la campaña de evaluación \HAHA. En este trabajo reportamos el desempeño del sistema para dos subtareas de la campaña: detección de humor y predicción de puntaje de diversión. También presentamos algunos problemas que surgen de la conversión entre los puntajes numéricos utilizados en los datos \HAHA y las anotaciones de juicio de pares de documentos requeridas para nuestro método.}}

\seplnkey{Computational humour, humour, Gaussian process preference learning, GPPL, best--worst scaling}

\seplnabstract{Most humour processing systems to date make at best discrete, coarse-grained distinctions between the comical and the conventional, yet such notions are better conceptualized as a broad spectrum.  In this paper, we present a probabilistic approach, a variant of Gaussian process preference learning~(GPPL), that learns to rank and rate the humorousness of short texts by exploiting human preference judgments and automatically sourced linguistic annotations.  We apply our system, which is similar to one that had previously shown good performance on English-language one-liners annotated with pairwise humorousness annotations, to the Spanish-language data set of the \HAHA evaluation campaign.  We report system performance for the campaign's two subtasks, humour detection and funniness score prediction, and discuss some issues arising from the conversion between the numeric scores used in the \HAHA data and the pairwise judgment annotations required for our method.}

\firstpageno{1}

\begin{document}



\label{firstpage} \maketitle\null\clearpage

\section{Introduction}
\label{sec:introduction}

Humour is an essential part of everyday communication, particularly in social media~\cite{holton2011journalists,shifman2013memes}, yet it remains a challenge for computational methods.  Unlike conventional language, humour requires complex linguistic and background knowledge to understand, which are difficult to integrate with NLP methods~\cite{hempelmann2008computational}.

An important step in the automatic processing of humour is to recognize its presence in a piece of text.  However, its intensity may be present or perceived to varying degrees to its human audience~\cite{bell2017failed}.  This level of appreciation (i.e., \emph{humorousness} or equivalently \emph{funniness}) can vary according to the text's content and structural features, such as nonsense or disparagement~\cite{carretero-dios2010assessing} or, in the case of puns, contextual coherence~\cite{lippman2000contextual} and the cognitive effort required to recover the target word~\cite[pp.\,123--124]{hempelmann2003paronomasic}.

While previous work has considered mainly binary classification approaches to humorousness, the \HAHA shared task~\cite{chiruzzo2019overview} also focuses on its gradation.  This latter task is important for downstream applications such as conversational agents or machine translation, which must choose the correct tone in response to humour, or find appropriate jokes and wordplay in a target language.  The degree of creativeness may also inform an application whether the semantics of a joke can be inferred from similar examples.

This paper describes a system designed to carry out both subtasks of the \HAHA evaluation campaign: binary classification of tweets as humorous or not humorous, and the quantification of humour in those tweets.  Our system employs a Bayesian approach---namely, a variant of Gaussian process preference learning~(GPPL) that infers humorousness scores or rankings on the basis of manually annotated pairwise preference judgments and automatically annotated linguistic features.  In the following sections, we describe and discuss the background and methodology of our system, our means of adapting the \HAHA data to work with our system, and the results of our system evaluation on this data.

\section{Background}\label{sec:related_work}

Pairwise comparisons can be used to infer rankings or ratings by assuming a \emph{random utility model}~\cite{thurstone1927law}, meaning that the annotator chooses an instance (from a pair or list of instances) with probability $p$, where $p$ is a function of the \emph{utility} of the instance.  Therefore, when instances in a pair have similar utilities, the annotator selects one with a probability close to 0.5, while for instances with very different utilities, the instance with higher utility will be chosen consistently.  The random utility model forms the core of two popular preference learning models, the Bradley--Terry model~\cite{bradley1952rank,luce1959possible,plackett1975analysis}, and the Thurstone--Mosteller model~\cite{thurstone1927law,mosteller1951remarks}.  Given this model and a set of pairwise annotations, probabilistic inference can be used to retrieve the latent utilities of the instances.

Besides pairwise comparisons, a random utility model is also employed by MaxDiff~\cite{marley2005some}, a model for best--worst scaling~(BWS), in which the annotator chooses the best and worst instances from a set.  While the term ``best--worst scaling'' originally applied to the data collection technique~\cite{finn1992determining}, it now also refers to models such as MaxDiff that describe how annotators make discrete choices.  Empirical work on BWS has shown that MaxDiff scores (instance utilities) can be inferred using either maximum likelihood or a simple counting procedure that produces linearly scaled approximations of the maximum likelihood scores~\cite{flynn2014best}.  The counting procedure defines the score for an instance as the fraction of times the instance was chosen as best, minus the fraction of times the instance was chosen as worst, out of all comparisons including that instance~\cite{kiritchenko2016capturing}.  From this point on, we refer to the counting procedure as BWS, and apply it to the tasks of inferring scores from  pairwise annotations for funniness.

Gaussian process preference learning (GPPL)~\cite{chu2005preference}, a Thurstone--Mosteller--based model that accounts for the features of the instances when inferring their scores, can make predictions for unlabelled instances and copes better with sparse pairwise labels.  GPPL uses Bayesian inference, which has been shown to cope better with sparse and noisy data~\cite{xiong2011bayesian,titov2012bayesian,beck2014joint,lampos2014predicting}, including disagreements between multiple annotators~\cite{cohn2013modelling,simpson2015language,felt2016semantic,kido2017}.  Through the random utility model, GPPL handles disagreements between annotators as noise, since no instance in a pair has a probability of one of being selected.

Given a set of pairwise labels, and the features of labelled instances, GPPL can estimate the posterior distribution over the utilities of any instances given their features.  Relationships between instances are modelled by a Gaussian process, which computes the covariance between instance utilities as a function of their features~\cite{rasmussen_gaussian_2006}.  Since typical methods for posterior inference~\cite{nickisch2008approximations} are not scalable (the computational complexity is $\mathcal{O}(n^3)$, where $n$ is the number of instances), we use a scalable method for GPPL that permits arbitrarily large numbers of instances and pairs~\cite{simpson2018finding}.  This method uses stochastic variational inference~\cite{hoffman2013stochastic}, which limits computational complexity by substituting the instances for a fixed number of \emph{inducing points} during inference.

The GPPL method has already been applied with good results to ranking arguments by convincingness (which, like funniness, is an abstract linguistic property that is hard to quantify directly) and to ranking English-language one-liners by humorousness~\cite{simpson2018finding,simpson2019predicting}.  In these two tasks, GPPL was found to outperform SVM and BiLSTM regression models that were trained directly on gold-standard scores, and to outperform BWS when given sparse training data, respectively.  We therefore elect to use GPPL on the Spanish-language Twitter data of the \HAHA shared task.

In the interests of replicability, we freely release the code for running our GPPL system, including the code for the data conversion and subsampling process detailed in §\ref{sec:data}.\footnote{\url{https://github.com/UKPLab/haha2019-GPPL}}

\section{Experiments}\label{sec:expts}

\subsection{Tasks}

The \HAHA evaluation campaign consists of two tasks.  Task~1 is humour detection, where the goal is to predict whether or not a given tweet is humorous, as determined by a gold standard of binary, human-sourced annotations.  Systems are scored on the basis of accuracy, precision, recall, and F-measure.  Task~2 is humorousness prediction, where the aim is to assign each funny tweet a score approximating the average funniness rating, on a five-point scale, assigned by a set of human annotators.  Here system performance is measured by root-mean-squared error~(RMSE).  For both tasks, the campaign organizers provide a collection of 24\,000 manually annotated training examples.  The test data consists of a further 6000 tweets whose gold-standard annotations were withheld from the participants.

\subsection{Data Preparation}\label{sec:data}

For each of the 24\,000 tweets in the \HAHA training data, the task organizers asked human annotators to indicate whether the tweet was humorous, and if so, how funny they found it on a scale from 1 (``not funny'') to 5 (``excellent'').  This is essentially the same annotation scheme used for the first version of the corpus~\cite{santiago2018crowd} which was used in the previous iteration of HAHA~\cite{castro2018overview}.  As originally distributed, the training data gives the text of each tweet along with the number of annotators who rated it as ``not humour'', ``1'', ``2'', ``3'', ``4'', and ``5''.  For the purposes of Task~1, tweets in the positive class received at least three numerical annotations and at least five annotations in total; tweets in the negative class received at least three ``not humour'' annotations, though possibly fewer than five annotations in total.  Only those tweets in the positive class are used in Task~2.

This ordinal data cannot be used as-is with our GPPL system, which requires as input a set of preference judgments between pairs of instances.  To work around this, we converted the data into a set of ordered pairs of tweets such that the first tweet has a lower average funniness score than the second.  (We consider instances in the negative class to have an average funniness score of 0.)  While an exhaustive set of pairings would contain 575\,976\,000 pairs (minus the pairs in which both tweets have the same score), we produced only 10\,730\,229 pairs, which was the minimal set necessary to accurately order the tweets.  For example, if the original data set contained three tweets $A$, $B$, and $C$ with average funniness scores 5.0, 3.0, and 1.0, respectively, then our data would contain the pairs $(C, B)$ and $(B, A)$ but not $(C, A)$.  To save memory and computation time in the training phase, we produced a random subsample such that the number of pairs where a given tweet appeared as the funnier one was capped at 500.  This resulted in a total of 485\,712 pairs.  In a second configuration, we subsampled up to 2500 pairs per tweet.  We used a random 60\% of this set to meet memory limitations, resulting in 686\,098 pairs.

With regards to the tweets' textual data, we do only basic tokenization as preprocessing.  For lookup purposes (synset lookup; see §\ref{sec:experimentalsetup}), we also lemmatize the tweets.

\subsection{Experimental Setup}\label{sec:experimentalsetup}

As we adapt an existing system that works on English data~\cite{simpson2019predicting}, we generally reuse the features employed there, but use Spanish resources instead.
Each tweet is represented by the vector resulting from a concatenation of the following:
\begin{itemize}
	\item The average of the word embedding vectors of the tweet's tokens, for which we use 200-dimensional pretrained Spanish Twitter embeddings~\cite{deriu2017embeddings}.
	\item The average frequency of the tweet's tokens, as determined by a Wikipedia dump.\footnote{\url{https://dumps.wikimedia.org/eswiki/20190420/eswiki-20190420-pages-articles.xml.bz2}; last accessed on 2019-06-15.}
	\item The average word polysemy---i.e., the number of synsets per lemma of the tweet's tokens, as given by the Multilingual Central Repository (MCR~3.0, release 2016)~\cite{gonzalezagirre2012mcr}.
\end{itemize}
Using the test data from the \HAHA[2018] task~\cite{castro2018overview} as a development set, we further identified the following features from the UO\_UPV system~\cite{ortegabueno2018haha} as helpful:
\begin{itemize}
	\item The heuristically estimated turn count (i.e., the number of tokens beginning with \texttt- or \texttt{-\null-}) and binary dialogue heuristic (i.e., whether the turn count is greater than 2).
	\item The number of hashtags (i.e., tokens beginning with \texttt\#).
	\item The number of URLs (i.e., tokens beginning with \texttt{www} or \texttt{http}).
	\item The number of emoticons.\footnote{\url{https://en.wikipedia.org/wiki/List_of_emoticons\#Western}, Western list; last accessed on 2019-06-15.}
	\item The character and token count, as well as mean token length.
	\item The counts of exclamation marks and other punctuation (\texttt{.,;?}).
\end{itemize}

We  adapt the existing GPPL implementation\footnote{\url{https://github.com/UKPLab/tacl2018-preference-convincing}} using 
the authors' recommended hyperparameter defaults~\cite{simpson2018finding}: 
batch size $|P_i|=200$, scale hyperparameters $\alpha_0=2$ and $\beta_0=200$,
and 
the number of inducing points (i.e., the smaller number of data points that act as substitutes for the tweets in the dataset)
$M=500$. The maximum number of iterations was set to 2000.
Using these feature vectors, hyperparameter settings, and data pairs, we require a training time of roughly two hours running on a 24-core cluster with 2\,GHz CPU cores.

After training the model, an additional step is necessary to transform the GPPL output values to the original funniness range (0, 1--5).
For this purpose, we train a Gaussian process regressor which we supply with the output values of the GPPL system as features and the corresponding \HAHA[2018] test data values as targets.
However, this model can still yield results outside the desired range when applied to the GPPL output of the \HAHA test data.
Thus, we afterwards map too-large and too-small values onto the range boundaries.
We further set an empirically determined threshold for binary funniness estimation.

\subsection{Results and Discussion}

\begin{table*}
  \centering
  \begin{tabular}{lccccc}
    \toprule
    & \multicolumn{4}{c}{Task 1} & Task 2 \\
    \cmidrule(lr){2-5}
    \cmidrule(lr){6-6}
    \multicolumn{1}{c}{System} & F\textsubscript{1} & Precision & Recall & Accuracy & RMSE\\
    \midrule
    \namecite{ismailov2019humor} & 0.821 & 0.791 & 0.852 & 0.855 & 0.736\\
    our system & 0.660 & 0.588 & 0.753 & 0.698 & 1.810 \\
    baseline & 0.440 & 0.394 & 0.497 & 0.505 & 2.455\\
    \bottomrule
  \end{tabular}
  \caption{Results for Task~1 (humour detection) and Task~2 (funniness score prediction)}
  \label{tab:task1}
\end{table*}


Table~\ref{tab:task1} reports results for the binary classification setup (Task~1) and the regression task (Task~2).  Included in each table are the scores of our own system, as well as those of the top-performing system~\cite{ismailov2019humor} and a naïve baseline.  For Task~1, the naïve baseline makes a random classification for each tweet (with uniform distribution over the two classes); for Task~2, it assigns a funniness score of 3.0 to each tweet.

In the binary classification setup, our system achieved an F-measure of 0.660 on the test data, representing a precision of 0.588 and a recall of 0.753.  In the regression task, we achieved RMSE of 1.810.  The results are based on the second data subsample (up to 2500 pairs), with the results for the first (up to 500 pairs) being slightly lower.  Our results for both tasks, while handily beating those of the naïve baseline, are significantly worse than those reported by some other systems in the evaluation campaign, including of course the winner.  This is somewhat surprising given GPPL's very good performance in previous English-language experiments~\cite{simpson2019predicting}.

\begin{figure}
	\centering
	\includegraphics[width=\linewidth]{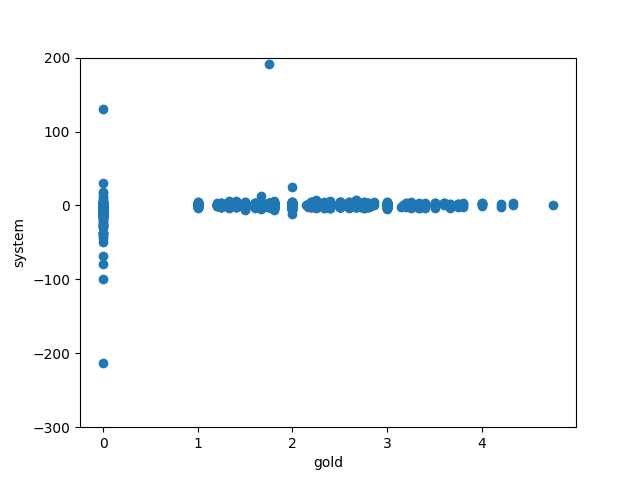}
	\caption{Gold values of the \HAHA[2018] test data ($x$ axis) vs.\ the scores assigned by our GPPL system ($y$ axis), before mapping to the expected funniness range using a Gaussian process regressor.  The lowest GPPL value (\textminus 1400) was removed from the plot to obtain a better visualization.}
	\label{haha2018gppl}
\end{figure}

Unfortunately, our lack of fluency in Spanish and lack of access to the gold-standard scores for the test set tweets precludes us from performing a detailed qualitative error analysis.  However, we speculate that our system's less than stellar performance can partly be attributed to the information loss in converting between the numeric scores used in the \HAHA tasks and the preference judgments used by our GPPL system.  In support of this explanation, we note that the output of our GPPL system is rather uniform; the scores occur in a narrow range with very few outliers.  (Figure~\ref{haha2018gppl} shows this outcome for the \HAHA[2018] test data.)  Possibly this effect would have been less pronounced had we used a much larger subsample, or even the entirety, of the possible training pairs, though as discussed in §\ref{sec:data}, technical and temporal limitations prevented us from doing so.  We also speculate that the Gaussian process regressor we used may not have been the best way of mapping our GPPL scores back onto the task's funniness scale (albeit still better than a linear mapping).

Apart from the difficulties posed by the differences in the annotation and scoring, our system may have been affected by the mismatch between the language resources for most of its features and the language of the test data.  That is, while we relied on language resources like Wikipedia and MCR that reflect standardized registers and prestige dialects, the \HAHA data is drawn from unedited social media, whose language is less formal, treats a different range of topics, and may reflect a wider range of dialects and writing styles.  Twitter data in particular is known to present problems for vanilla NLP systems, at least without extensive cleaning and normalization~\cite{ws-2015-noisy}.
This is reflected in our choice of word embeddings: while we achieved a Spearman rank correlation of $\rho = $ 0.52 with the \HAHA[2018] test data using embeddings based on Twitter data~\cite{deriu2017embeddings}, the same system using more ``standard'' Wikipedia-\slash news-\slash Web-based embeddings\footnote{\url{https://zenodo.org/record/1410403}} resulted in a correlation near zero.

\section{Conclusion}
\label{sec:conclusion}

This paper has presented a system for predicting both binary and graded humorousness.  It employs Gaussian process preference learning, a Bayesian system that learns to rank and rate instances by exploiting pairwise preference judgments.  By providing additional feature data (in our case, shallow linguistic features), the method can learn to predict scores for previously unseen items.

Though our system is based on one that had previously achieved good results with rudimentary, task-agnostic linguistic features on two English-language tasks (including one involving the gradation of humorousness), its performance on the Spanish-language Twitter data of \HAHA was less impressive.  We tentatively attribute this to the information loss involved in the (admittedly artificial) conversion between the numeric annotations used in the task and the preference judgments required as input to our method, and to the fact that we do not normalize the Twitter data to match our linguistic resources.

A possible avenue of future work, therefore, might be to mitigate the data conversion problem.  However, as it has been rather convincingly argued, both generally~\cite{thurstone1927law} and in the specific case of humour assessment~\cite{shahaf2015inside}, that aggregate ordinal rating data should not be treated as interval data, the proper solution here would be to recast the entire task from one of binary classification or regression to one of comparison or ranking.  Perhaps the best way of doing this would be to source new gold-standard preference judgments on the data, though this would be an expensive and time-consuming endeavour.\footnote{We note with appreciation that the upcoming SemEval-2020 task on humour assessment~\cite{hossain2020semeval} does include a subtask for predicting preference judgments, though it seems the underlying gold-standard data still uses aggregate ordinal data.}

Regardless of the task setup, there are a few further ways our system might be improved.  First, we might try normalizing the language of the tweets, and secondly, we might try using additional, humour-specific features, including some of those used in past work as well as those inspired by the prevailing linguistic theories of humour~\cite{attardo1994linguistic}.  The benefits of including word frequency also point to possible further improvements using $n$-grams, TF--IDF, or other task-agnostic linguistic features.

\subsection*{Acknowledgments}

The authors thank Gisela Vallejo for her help with Spanish-language translations.

This work has been supported by the German Federal Ministry of Education and Research (BMBF) under the promotional reference 01UG1816B (CEDIFOR), by the German Research Foundation (DFG) as part of the QA-EduInf project (grants GU\,798\slash \mbox{18-1} and RI\,803/12-1), by the DFG-funded research training group ``Adaptive Preparation of Information from Heterogeneous Sources'' (AIPHES; GRK\,1994/1), and by the Austrian Science Fund (FWF) under project M\,2625-N31.  The Austrian Research Institute for Artificial Intelligence is supported by the Austrian Federal Ministry for Science, Research and Economy.

\bibliographystyle{fullname}
\bibliography{2020_Miller_SEPLN}

\end{document}